\documentclass[oneside,11pt, a4paper, footinclude=true, headinclude=true, cleardoublepage=empty]{scrbook}
\usepackage[linedheaders,parts,pdfspacing,dottedtoc]{classicthesis}
\usepackage{amsmath}
\usepackage{acronym}
\usepackage[a4paper, hmargin={2.8cm, 2.8cm}, vmargin={2.5cm, 2.5cm}]{geometry}
\usepackage{eso-pic} 
\usepackage{graphicx} 
\usepackage[utf8]{inputenc}
\usepackage[english]{babel}
\usepackage{csquotes}
\usepackage{bookmark}
\usepackage{longtable}
\usepackage{array}
\usepackage{multirow}
\usepackage{times}
\usepackage{lingmacros}
\usepackage{color, colortbl}
\usepackage{tabularx}
\usepackage{pdfpages}
\usepackage{footnote}
\usepackage{booktabs}
\usepackage{pdfpages}
\usepackage{blindtext}

\DeclareUnicodeCharacter{0301}{\'{e}}

\definecolor{mygray}{rgb}{0.86,0.86,0.86}
\makesavenoteenv{tabular}
\makesavenoteenv{table}

\def\signed #1{{\leavevmode\unskip\nobreak\hfil\penalty50\hskip2em
  \hbox{}\nobreak\hfil(#1)%
    \parfillskip=0pt \finalhyphendemerits=0 \endgraf}}

    \newsavebox\mybox

\usepackage[backend=biber,style=authoryear,dashed=false,sorting=nty, maxcitenames=2]{biblatex}

\def \ColourPDF {Images/ku-farve.pdf}
\def \TitlePDF   {Images/ku-en.pdf}

\subject{ 
  \vspace{3.5cm}
  \Large{PhD thesis in Department of Computer Science } \\}

\title{
  \Huge{Semantic Representation and Inference for NLP}\\
}

\author{
  \Large{Dongsheng Wang} \\
  \texttt{wang@di.ku.dk}
  \vspace{1.5cm} \\
  \large{Supervisor: Christina Lioma and Jakob Grue Simonsen}
}
\date{September 2020}

\begin{document} 
\pagenumbering{roman}
\AddToShipoutPicture*{\put(0,0){\includegraphics*[viewport=0 0 700 600]{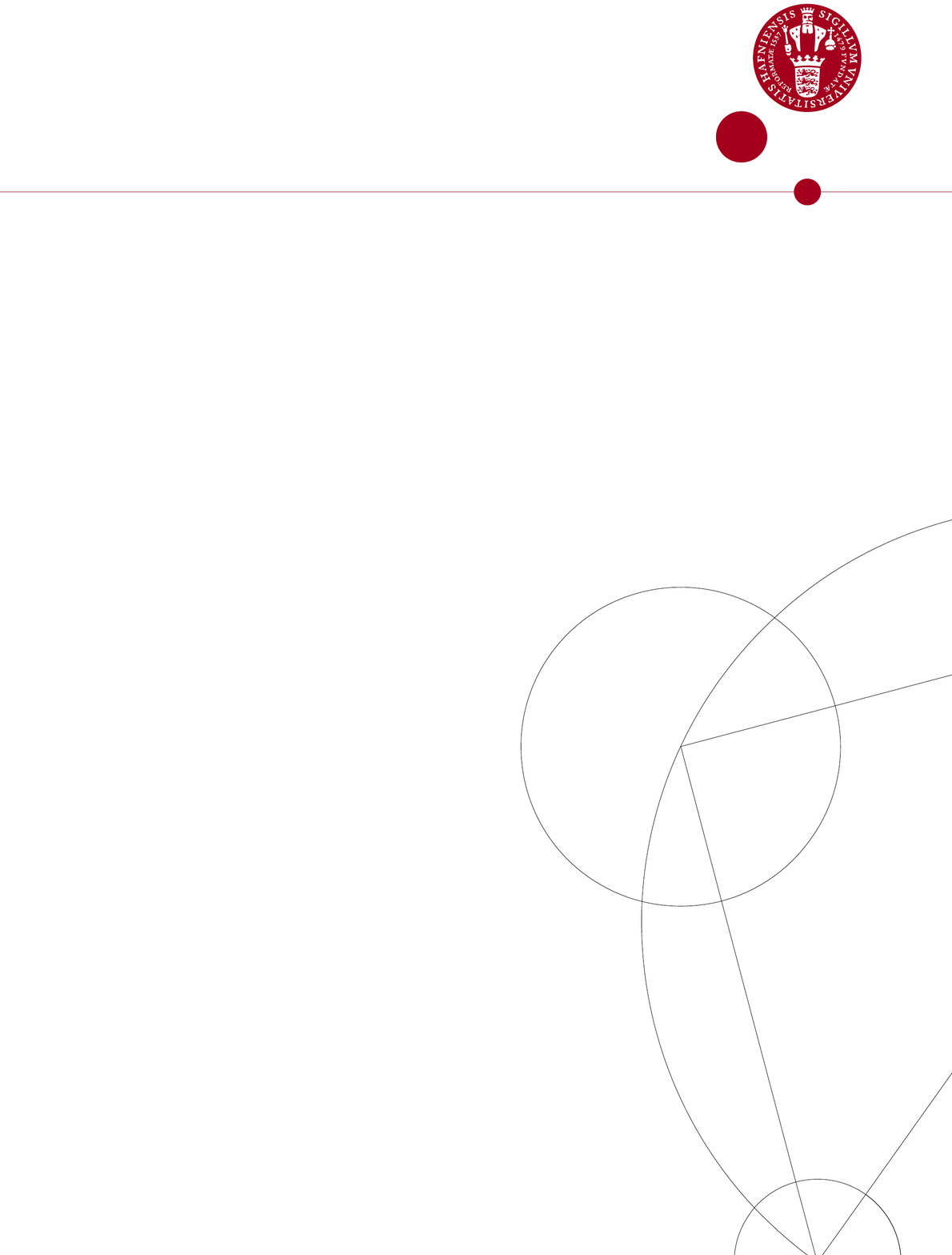}}}
\AddToShipoutPicture*{\put(0,602){\includegraphics*[viewport=0 600 700 1600]{\ColourPDF}}}
\AddToShipoutPicture*{\put(0,0){\includegraphics*{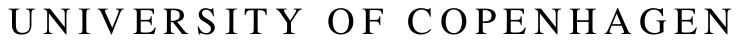}}}

\clearpage\maketitle
\thispagestyle{empty}

\newpage

\chapter*{Acknowledgements}

First of all, I thank my supervisors Christina and Jakob, for being my supervisors. I would not have walked through my PhD and have this thesis written without them. Especially, I thank Christina for the thorough guidance from all aspects; she is a very responsible and respectful professor. She is highly professional and scientific in researching, teaching, and management, as our supervisor in the IR lab and as the head of the machine learning section. It also has been great to have Jakob as our secondary supervisor; Jakob is both a humorous and gentlemanly professor bringing us much fun and help. He was recently promoted to be the head of the computer science department, which did not surprise me because I have always been surprised by his broad knowledge and interpersonal capacity to invent a peaceful environment. I also thank Peter Bruza from QUT (Australia) and Ingo Schmitt from BTU (Germany), who had accommodated my academic visiting in their labs for three months, respectively, and I spent pleasant time both living in their cities and collaborating with them. 

It is hard to forget the first Scandinavian winter that I experienced in Denmark - cold and dark. It could have been tough for me if I had not met so many lovely and friendly colleagues in the machine learning section at DIKU who came to me and took me for a drink and talk. I also thank all the young radical ESRs in the QUARTZ project. We had a pleasant time visiting each other across Europe and the UK to share the talks, ideas and collaborate. I want to thank IR lab colleagues Lucas, Casper, Christian, and later on, joined Maria. We sit together, work, talking, and collaborating closely in the shared office. 

Finally, my Ph.D. is funded by the European Union Project - QUARTZ (Quantum Information Access and Retrieval Theory) \footnote{https://www.quartz-itn.eu/}. I thank EU Marie-Curie for funding this excellent research project.

\chapter*{Abstract}

Semantic representation and inference is essential for Natural Language Processing (NLP). The state of the art for semantic representation and inference is deep learning, and particularly Recurrent Neural Networks (RNNs), Convolutional Neural Networks (CNNs), and transformer Self-Attention models. This thesis investigates the use of deep learning for novel semantic representation and inference, and makes contributions in the following three areas: creating training data, improving semantic representations and extending inference learning.
In terms of creating training data, we contribute the largest publicly available dataset of real-life factual claims for the purpose of automatic claim verification (MultiFC), and we present a novel inference model composed of multi-scale CNNs with different kernel sizes that learn from external sources to infer fact checking labels. In terms of improving semantic representations, we contribute a novel model that captures non-compositional semantic indicators. By definition, the meaning of a non-compositional phrase cannot be inferred from the individual meanings of its composing words (e.g., hot dog). Motivated by this, we operationalize the compositionality of a phrase contextually by enriching the phrase representation with external word embeddings and knowledge graphs. Finally, in terms of inference learning, we propose a series of novel deep learning architectures that improve inference by using syntactic dependencies, by ensembling role guided attention heads, incorporating gating layers, and concatenating multiple heads in novel and effective ways.
This thesis consists of seven publications (five published and two under review).

\chapter*{Resumé}

Semantisk repræsentation og inferens er essentielt for Natural Language Processing (NLP). State-of-the-art indenfor semantisk repræsentation og inferens er deep learning, og særligt recurrent neural networks (RNNs), convolutional neural networks (CNNs), og transformer Self-Attention modeller. Denne afhandling undersøger brugen af deep learning for nye metoder indenfor semantisk repræsentation og inferens, og laver videnskabelige bidrag indenfor de følgende områder: lave trænings data, forbedre semantisk repræsentation, og udvide inferenslæring. Med hensyn til at lave træningsdata, bidrager vi med det største tilgængelige datasæt af faktuelle udsagn med det formål at lære at automatisk verificere disse udsagn (MultiFC), og vi præsenterer en ny inferensmodel bestående af multi-scale CNNs med forskellige kernel størrelser, som lærer fra eksterne kilder at inferere fakta-checking labels. I forbindelse med at forbedre semantisk repræsentationer, bidrager vi med en ny model der inkorporerer ikke-kompositionelle semantiske indikatorer. Per definition, kan betydningen af en ikke-kompositionssætning ikke udledes af de enkelte betydninger af dets enkelte ord (f.eks. hot dog). Motiveret af dette operationaliserer vi kompositionaliteten af en sætning kontekstmæssigt ved at berige sætningerepræsentation med word embeddings og knowledge graphs. Endelig med hensyn til inferenslæring bidrager vi med en række nye deep learning arkitekturer, der forbedrer inferens ved hjælp af syntaktiske afhængigheder ved at samle rollestyret attention hoveder, ved at inkorporerer gating lag, og sammenkæde flere hoveder på nye og effektive måder. Denne afhandling består af syv publikationer (fem udgivet og to under gennemgang).

\pdfbookmark[1]{\contentsname}{tableofcontents}

\setcounter{tocdepth}{2} 
\setcounter{secnumdepth}{3} 

\tableofcontents 


\pdfbookmark[1]{\listfigurename}{lof}
\listoffigures

\pdfbookmark[1]{\listtablename}{lot}
\listoftables
  
   

\pagenumbering{arabic}

\part{Comprehensive Summary}\label{part:I}

\chapter{Roadmap}

The thesis is structured in five parts, composed of a total of 13 chapters. The first part is the comprehensive summary, which summarizes the backgrounds, objectives, contributions, and conclusions. From the second part onward, we include the compilation of the original research papers.

Part I is the comprehensive summary, consisting of the first six chapters. In chapter \ref{c:background}, we present the necessary theoretical background and build the intuition for understanding the remaining of the thesis. Precisely, we introduce semantic representation and inference learning for Natural Language Processing (NLP), especially the state of the art approaches of deep learning such as Recurrent Neural Networks (RNNs), Convolutional Neural Networks (CNNs), and Transformer Self-Attention models. In chapter \ref{c:objective}, we detail the research questions and our objectives, derived from the included papers. Briefly, we investigate deep learning usage for novel semantic representation and inference learning and make contributions in the following three primary objectives: creating training data, improving semantic representation, and extending inference learning. The main contributions and results of the thesis are summarized in chapter \ref{c:contribution}, followed by future work and conclusion in chapter \ref{c:futurewrk} and \ref{c:conclusion}.

Part II to part V are the compilation of original research papers. Specifically, chapter \ref{chap:createdataset} introduces the work about creating a dataset; chapter \ref{chap:comp_detection} introduces the novel model for compositionality detection for semantic representation; chapters \ref{chap:fact_checking} and \ref{chap:structblock} describe two studies on extending convolutional neural-based inference learning. Chapter \ref{chap:guidedheads}, \ref{chap:gatedheads}, and \ref{chap:msencoder} introduce three methods for extending self-attention based inference learning.


\newcommand{\softmaxname}[0]{\ensuremath{\textrm{softmax}}}

\chapter{Background}
\label{c:background}
\section{Semantic Representation}
Semantic representation of words and phrases in a machine-interpretable form has been an important NLP goal. We now overview the word embedding and compositionality detection techniques, which form the theoretical background that this thesis extends.


\subsection{Word Embedding}
One-hot encoding technique is a common and basic way of turning a word into a vector. One-hot encoding generates a binary vector of size $N$ (the size of the vocabulary) where each word corresponds to a unique integer index. As a result, a word can be expressed with a vector with all zeros except for the word's index entry, which is one. However, the disadvantage of the one-hot encoding is that it can lead to a sparse, high dimensional vector from a large number of words in the vocabulary. 

Contrarily, word embeddings are dense (low dimensional) vectors. Word embedding, also known as the distributed representation of words (\cite{mikolov2013distributed}), refers to a set of machine learning algorithms that learn real-valued dense vector representation for each vocabulary term in a corpus. 

\begin{figure}
    \centering
    \includegraphics[scale=0.9]{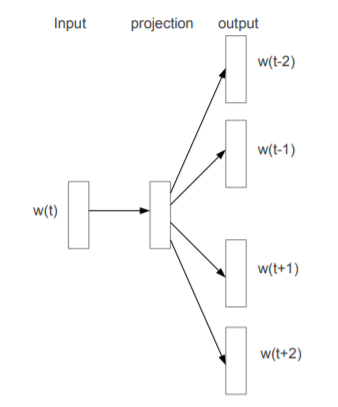}
    \caption{The Skip-gram model, designed to learn word vector representations that are good at predicting nearby terms. (\cite{mikolov2013distributed}).}
    \label{fig:skipgram}
\end{figure}

A popular model for learning word embedding is neural network-based language models, e.g., the word2vec model (\cite{mikolov2013distributed}). word2vec is an embedding model that learns word vectors via a neural network with a single hidden layer. Two implementations of the model include the continuous bag of words (CBOW) and the skip-gram, as shown in Figure \ref{fig:skipgram}. CBOW predicts a target word with its context words while skip-gram, conversely, predicts the context words from its target word. Another widely used model for learning word embedding is the Global vectors (GloVe) (\cite{pennington2014glove}). GloVe employs global matrix factorization over word-word matrices.

Word embeddings have been successfully employed in many NLP and Information Retrieval (IR) tasks. For example, word embedding vectors are applied in information retrieval (\cite{ganguly2015word}), recommendation systems (\cite{ozsoy2016word}), text classification (\cite{ge2017improving}), etc.

It is worth to mention that a more recent and more powerful embedding technique is the BERT (Bidirectional Encoder Representations from Transformers) (\cite{devlin2018bert}), released in 2018.  BERT's core component is the Transformer's encoder representation. BERT practically pre-trains the bidirectional encoder representation from the Transformer (which is introduced in chapter \ref{ss:transformer}) on unlabelled texts with masks; therefore, BERT is also called the masked language model. 

\subsection{Compositionality detection}
Compositionality in NLP describes the extent to which the meaning of a phrase can be decomposed into the meaning of its individual components and the way these components are combined. Compositionality plays an vital role in word embeddings because a non-decomposable phrase should, in principle, be treated as a single semantic unit instead of a bag of word (BOW) in word embedding approaches.  For example, brown dog is a fully compositional phrase meaning a dog of brown color whereas hot dog is a non-compositional phrase denoting a type of food.

Automatic compositionality detection has received attention for almost two decades. Earlier approaches mostly measure the similarity between the original phrase and its component words. For instance, Baldwin et al. and Katz et al. (\cite{baldwin2003empirical,katz2006automatic}) utilize Latent Semantic Analysis (LSA) to calculate the semantic similarity and hence to measure compositionality. Venkatapathy et al. (\cite{venkatapathy2005measuring}) extend this approach by incorporating collocation features, e.g., phrase frequency, point-wise mutual information, extracted from the British National Corpus. 

Another line of work computes the similarity between a phrase and the phrase's perturbed versions where the words are replaced, one at a time, by their synonyms. For instance, Kiela et al. (\cite{Kiela13}) calculate the semantic distance between a phrase and its perturbation, with cosine similarity, which computes a phrase weight by pointwise-multiplication vectors of its terms. Lioma et al. (\cite{lioma2015non}) calculate the semantic distance with Kullback-Leibler divergence utilizing a language model; and, in subsequent work, Lioma et al. (\cite{Lioma2017}) represent the original phrase and its perturbations as ranked lists, which are used to measure their correlation or distance. As a result, the compositionality score $score(p)$ of the phrase $p$ can be expressed as follows,
\begin{equation}
    score(p) = \frac{\sum_{\hat{p} \in S(p)} sim(\mathcal{R}(p),\mathcal{R}(\hat{p}))}{|S(p)|}
\end{equation}
where $S(p)$ indicates the perturbation set where $\hat{p}$ is its element; $\mathcal{R}(\cdot)$ is the semantic representation of the phrase (e.g., word embedding, ranked list, etc.).      

Another line of work uses word embeddings and deep artificial neural networks for compositionality detection. Salehi et al. (\cite{salehi2015word}) employ the word-based skip-gram model for learning non-compositional phrases, treating phrases as individual tokens with vectorial composition functions. Hashimoto and Tsuruoka (\cite{hashimoto2016adaptive}) employ syntactic features, including word index, frequency, and PMI of a phrase and its components words to learn the embeddings. Yazdani et al. (\cite{yazdani2015learning}) utilize a polynomial projection function and deep artificial neural networks to learn the semantic composition and detect non-compositional phrases like those that stand out as outliers, assuming that the majority are compositional.

One of this thesis's contributions is a new compositionality model that extends the above state of the art. We discuss this in section \ref{s:semrepre}.


\section{Inference Learning approaches}

We now overview the major deep learning architectures that are commonly used to make semantic inferences in text, which forms the theoretical background that this thesis extends.

\subsection{Recurrent Neural Network}
Artificial neural networks (ANNs), usually called neural networks (NNs), are computing systems inspired by the biological neural networks of the human brain (\cite{demuth2014neural}). A neural network contains layers of interconnected nodes where each node is a mathematical function used to translate a data input of one form into the desired output of another form. Neural networks are one of the approaches in machine learning algorithms.

RNN (\cite{schuster1997bidirectional}) is a type of artificial neural network where the connections between nodes make a directed graph along a temporal sequence. This architecture is designed to capture temporal dynamic behavior. From the perspectives of NLP, temporal behavior corresponds to word positions in a sentence. To give a formal definition of RNNs, we assume $x = (x_1, x_2, x_3,..., x_T)$ represents a sequence of $T$ words, and $h_t$ represents RNN memory at time step $t$, an RNN model updates its memory information using:

\begin{equation}
\label{eq:vanilla_rnn}
h_t = \sigma (W_x x_t + W_h h_{t-1} + b_t)
\end{equation}
where $\sigma$ is a nonlinear activation function (e.g., logistic sigmoid or rectified linear unit (ReLU)), $x_t$ is the word at position $t$, $h_t$ represents RNN memory at time step $t$;  $W_x$ and $W_h$ are weight matrices of the input and the recurrent connections respectively, which are learned in neural model; and $b_t$ is a constant bias vector which also need to be learned. Eq. \ref{eq:vanilla_rnn} is a vanilla RNN model, presenting a transition from one state to another, rather than variable-length states. Therefore, the learned model has the advantage of being applicable to variable sequence length input.

However, the vanilla RNN model's drawback is that the stochastic gradient descent may blow up or vanish over time caused by error signals flowing backward, called "vanishing gradients" and "exploding gradients". Precisely, in a neural network of $n$ hidden layers, "vanishing gradients" indicates when the derivatives are small, these $n$ derivatives are multiplied together, and the gradient decrease exponentially as we propagate down until vanishing. Alternatively, if the derivatives are large, the gradient will increase exponentially as we propagate down the model resulting in very large updates, which is called the "exploding gradients" problem.

These problems make RNNs hard to train. Therefore, the extensions of RNN architectures have been developed to overcome these problems, including long short term memory (LSTM) (\cite{hochreiter1997long}) and Bidirectional LSTM (BiLSTM) (\cite{bin2016bidirectional}). 

The LSTM models extend the RNNs’ memory in order to keep and learn long-term dependencies of inputs. The LSTM memory is called a “gated” cell, which is functioned to make the decision of preserving or ignoring the memory information. Precisely, an LSTM model consists of three gates, forget, input, and output gates. The forget gate makes the decision to preserve or remove the existing information; the input gate controls the extent to which the new information should be added into the memory. The output gate specifies what part of the LSTM memory contributes to the output. BiLSTM is an extension of LSTM that merges two LSTMs traversing the input twice 1) left-to-right and 2) right-to-left).

LSTM has been reported superior performance compared to a more traditional time series model, Autoregressive Integrated Moving Average (ARIMA), with a large margin (\cite{siami2019performance}). ARIMA model is used to fits the non-stationary time series, which effectively transforms the non-stationary data into a stationary one. In addition, BiLSTM has been demonstrated improved performance than LSTM in time series prediction, though BiLSTM is more time-consuming than LSTM.

\subsection{Convolutional Neural Networks}
CNNs are originally proposed for computer vision and have achieved remarkable results for image classification (\cite{krizhevsky2012imagenet}). Consequently, Kim et al. (\cite{kim2014convolutional}) utilize the CNN variant on sentence classification task and demonstrate the effectiveness of CNNs on NLP as well. Afterward, the CNNs are employed and implemented on NLP tasks by many other studies (\cite{jacovi2018understanding,bai2018empirical,johnson2014effective}).

\begin{figure}
    \centering
    \includegraphics[scale=0.7]{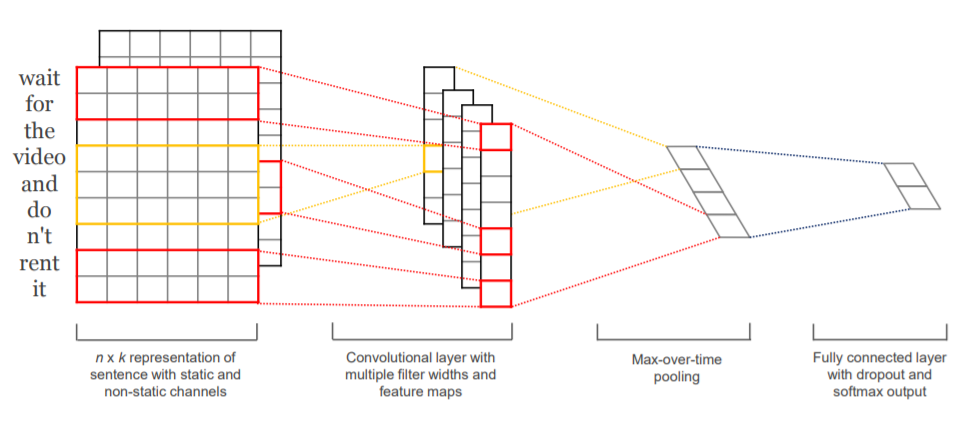}
    \caption{An example of CNNs with two channels (\cite{kim2014convolutional}).}
    \label{fig:cnn_arch}
\end{figure}

We introduce the CNN model on NLP as shown in figure \ref{fig:cnn_arch}. Given $x_i \in R^d$ the d-dimensional word vector corresponding to the i-th word in the sentence. A sentence of length n is represented as $x_{(1:n)} = x_1 \bigoplus x_2 \bigoplus ... x_n $, where $\bigoplus$ is the concatenation operator. In general, let $x_{i:i+j}$ refer to the concatenation of words $x_i, x_{i+1}, . . . , x_{i+j}$. A convolution operation involves a filter $w$, which is applied to a window of $h$ words to generate a new feature. A feature $c_i$ that is generated from a window of words $x_{i:i+h-1}$ is expressed as,
\begin{equation}
    c_i = \sigma(w \cdot x_{i:i+h-1} + b)
\end{equation}

where $\sigma$ is the non-linear activation function (e.g. hyperbolic tangent); a filter is defined as $w \in R
^{hd}$, where $d$ is the word embedding dimension and is applied to a
window of $h$ words. This filter is applied to all possible windows of $\{x_{1:h}, x_{2:h+1},...,x_{n-h+1:n}\}$ in the sentence and produce features as below,
\begin{equation}
    c = [c_1,c_2, ... c_{n-h+1}]
\end{equation}
Over the feature map, we take the maximum values $\hat{c} = max\{c\}$ as the feature value for this specific filter $w$.

One of this thesis's contributions is an extension of the convolutional neural model that improves inference effectiveness. We discuss this in section \ref{s:extendcnn}.

\subsection{Transformer}
\label{ss:transformer}
The Transformer (\cite{vaswani2017attention}) is initially proposed as a sequence-to-sequence (seq2seq) model but has also been used successfully for transfer learning tasks, especially after being pre-trained on massive amounts of unlabeled data. seq2seq indicates the input takes a sequence of items (words, letters, time series, etc.), and the output gives another sequence of items.

\begin{figure}
    \centering
    \includegraphics[scale=0.8]{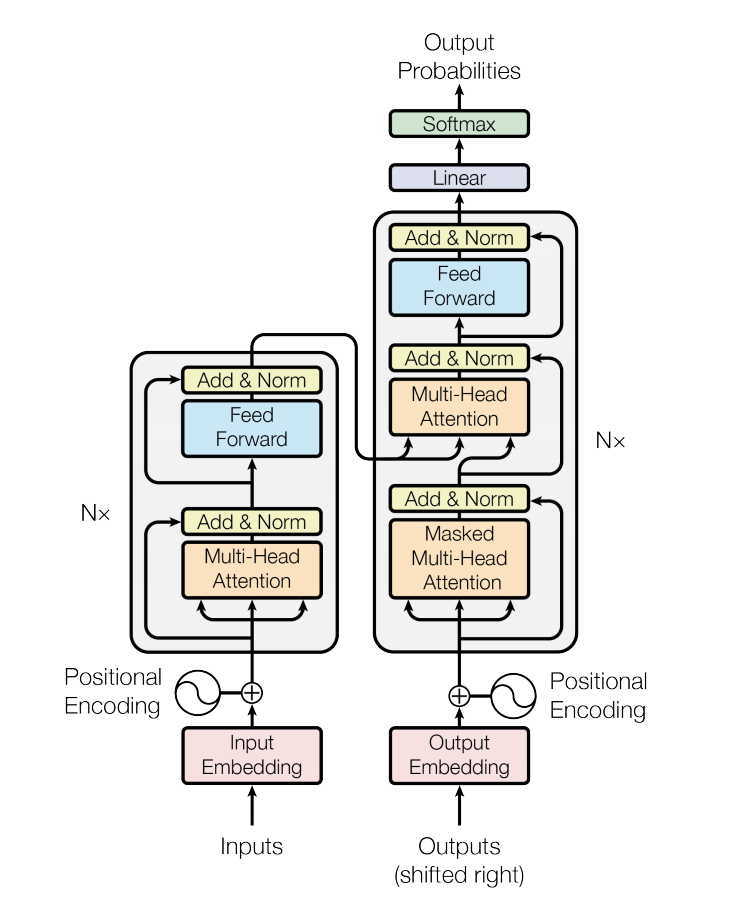}
    \caption{Architecture of vanilla Transformer (\cite{vaswani2017attention})}
    \label{fig:transformer_arc}
\end{figure}

As shown in Figure \ref{fig:transformer_arc}, the model is composed of an encoder (left side) and a decoder (right side). The encoder is often employed independently for transfer learning tasks. In general, the essential part is the multi-head attention component, for which a single attention head is computed with:
\begin{equation}
    \label{eq:attent}
    \textrm{Attention}(Q,K,V)=\softmaxname(\frac{QK^T}{\sqrt{d_k}}) V
\end{equation}
\noindent where $Q$ is the query, $K$ is the key, $V$ is the value, and $d_k$ is the key dimension. The input to each head is a head-specific linear projection, and the attention for each head is concatenated for a single output. The number of layers is a parameter that can be tuned. 

One of this thesis's contributions is an extension of the self-attention model that improves the above vanilla self-attention model. We discuss this in section \ref{s:extendsa}.

\section{Inference Tasks}
This section briefly summarizes the inference tasks that are involved in this thesis. The tasks include text classification, recognizing text entailment, paraphrasing, fact-checking, relation extraction, and machine translation.

\subsection{Text Classification}
Given some texts as input, the task of text classification is to identify which class each text belongs to. The classes are predefined, which can be binary (two classes) or multi-class (three or more classes). Classification (including sentiment and topic classification) is one of the most typical tasks in NLP. Two examples are shown in Table \ref{tb:class_examps}.

\begin{table}[]
\centering
\small
\caption{\label{tb:class_examps} Examples for text classification task.}
\begin{tabular}{@{}lll@{}}
\toprule
type & input & output \\ \midrule
Sentiment & It was a nice trip! & Positive \\
Topical & My investment in stock brought me large profit. & Finance \\ \bottomrule
\end{tabular}
\end{table}


\subsection{Recognizing Text Entailment}
Given some pairs of texts as input, the task of recognizing text entailment is to identify whether the semantic meaning of one text is entailed or can be inferred from another text. The output is binary (True or False). An example is shown in Table \ref{tb:rte_examps}. 

\begin{table}[]
\centering
\small
\caption{\label{tb:rte_examps} An examples for recognizing text entailment task.}
\begin{tabular}{@{}llll@{}}
\toprule
ID & Sentence1 & Sentence 2 & output \\ \midrule
01 & The engine stoped all of a sudden & The motor cut out abrubtly & True \\ \bottomrule
\end{tabular}
\end{table}

\subsection{Paraphrasing Task}
Given some pairs of texts as input, the task of paraphrasing is to recognize whether each pair of texts captures a paraphrase/semantic equivalence relationship. A paraphrase is the restatement of the meaning of a text using different words. The output is binary (paraphrase or non-paraphrase). An example is shown in Table \ref{tb:paraphrase_examps}.

\begin{table}[]
\centering
\small
\caption{\label{tb:paraphrase_examps} An examples for Paraphrasing task.}
\begin{tabular}{lll}
\hline
 & Pairs of sentences & Output \\ \hline
\multicolumn{1}{l}{Sentence 1} & \multicolumn{1}{l}{I would rather be talking about positive numbers than negative.} & \multicolumn{1}{l}{\multirow{2}{*}{paraphrase}} \\ \cline{1-2}
Sentence 2 & \multicolumn{1}{l}{But I would rather be talking about high standards rather than low standards} & \multicolumn{1}{l}{} \\ \hline
\end{tabular}
\end{table}


\subsection{Fact-checking}
Given some claims as input, the task of fact-checking is to check the factuality of these claims. Factuality indicates the degree of being actual in terms of right or wrong. The predefined output can be binary (e.g., true or false) or multi-class (e.g., true, false, half-true,  mostly-true, etc.). One example is shown in the Table \ref{tb:factchecking_exam}. fact-checking is receiving extraordinary attention in recent years. 

\begin{table}[]
\centering
\small
\caption{\label{tb:factchecking_exam} An example of fact-checking.}
\begin{tabular}{@{}lll@{}}
\toprule
Speaker & Sentence (Claim) & output \\ \midrule
Trump & Biden's plan is a 14\% tax hike on middle class families. & False \\ \bottomrule
\end{tabular}
\end{table}

\subsection{Relation Extraction}
Given some texts and entities from the texts as input, the task of relation extraction is to identify the semantic relations between two or more entities. The semantic relations are predefined, considering direction. The direction of relation means who modifies what. An example is shown in Table \ref{tb:relationextraction}, which can be expressed using the triple format (Steve Jobs, Founder, Apple) rather than (Apple, Founder, Steve Jobs).

\begin{table}[]
\centering
\footnotesize
\caption{\label{tb:relationextraction} An example of relation extraction. Entities are marked with red color.}
\begin{tabular}{@{}ll@{}}
\toprule
Sentence & Relation \\ \midrule
\textbf{\color{red} Steve Jobs} and Wozniak co-founded \textbf{\color{red} Apple} in 1977, introducing first the Apple I and then the Apple II. & Founder \\ \bottomrule
\end{tabular}
\end{table}

\subsection{Machine Translation Task}
Given some texts as input, the task of machine translation is to automatically convert the natural language into another natural language with the meaning of the input text preserved, producing fluent text in the output language. An example of a machine translation task from English to German is shown in Table \ref{tb:mttask}.

\begin{table}[]
\centering
\footnotesize
\caption{\label{tb:mttask} An example of machine translation task.}
\begin{tabular}{@{}ll@{}}
\toprule
Input (English text) & output (German text) \\ \midrule
A man in an orange hat staring at something. & Ein Mann mit einem orangefarbenen Hut, der etwas anstarrt. \\ \bottomrule
\end{tabular}
\end{table}

\section{Evaluation Metrics}
\subsection{Classification Metrics}
To introduce Accuracy, Precision, Recall, F measure evaluation metrics, we first introduce the confusion matrix (or contingency table) for binary classification, as shown in Table \ref{tb:confutionmatrix}. The terms of positive and negative refer to the classifier's prediction, and the terms true and false indicates whether that prediction corresponds to the external judgment.

\begin{table}[]
\caption{\label{tb:confutionmatrix} Confusion matrix for binary classification.}
\centering
\small
\begin{tabular}{@{}lll@{}}
\toprule
 & \textbf{Actual positive} & \textbf{Actual negative} \\ \midrule
\textbf{Predicted positive} & True positive (tp) & False positive (fp) \\
\textbf{Predicted negative} & False negative (fn) & True negative (tn) \\ \bottomrule
\end{tabular}
\end{table}

\begin{itemize}
    \item \textbf{Accuracy (ACC)}: the proportion of correct predictions (both true positives and true negatives) to the total number of predictions. It is defined as Eq. \ref{eq:accu}
\begin{equation}
\label{eq:accu}
    Accuracy = \frac{tp + tn}{tp+tn+fp+fn}
\end{equation}
    \item \textbf{Precision (P)}: the proportion of correct positive predictions (true positive) to all the positive predictions (true positive and false positive). It is defined as Eq. \ref{eq:precision},
\begin{equation}
\label{eq:precision}
Precision = \frac{tp}{tp+fp}
\end{equation}
    \item \textbf{Recall (R)}: the proportion of correct positive predictions (true positive) to all the positive samples.
\begin{equation}
\label{eq:recall}
Recall = \frac{tp}{tp+fn}
\end{equation}
    \item \textbf{F1 Score}: the harmonic mean of precision and recall. 
\begin{equation}
\label{eq:f1}
F1 = \frac{2P\times R}{P+R}
\end{equation}    
\end{itemize}

For multi-class classification, the confusion matrix can be considered as a detailed breakdown of correct and incorrect classifications for each class.

These classification metrics can be applied to evaluate recognizing text entailment, paraphrasing, fact-checking, and relation extraction tasks as well.

\subsection{Machine Translation Metrics}
We evaluate the machine translation performance using the Bilingual Evaluation Understudy (BLEU) measure (\cite{papineni2002bleu}). BLEU is used to compute the geometric average of the test corpus' modified precision scores and then multiply the result by an exponential brevity penalty factor. The BLEU is expressed as,
\begin{equation}
    BLEU = BP \cdot exp(\sum_{n=1}^{N} w_n log p_n )
\end{equation}

where BP means brevity penalty with the value $BP=1 if c>r$ or $BP = e^{(1-r)/c}$; $c$ is the length of the candidate translation and $r$ is the effective reference corpus length. The modified n-gram precision is expressed as $p_n$, using n-grams up to length $N$; and the positive weights $w_n$ sum up to one.

\subsection{Data Splitting and Cross Validation}
Data splitting and re-sampling are fundamental techniques to build reliable prediction models. Model performance metrics evaluated using training samples (also called in-sample) are retrodictive, not predictive. Without evaluating models with future samples, the models can be overfitting (learning too much from the data) or underfitting (learning too little from the data).

\textbf{Data splitting}. It is common to split data into multiple parts: training, validation, and test sets. The training set is used to design the models; the validation set is used to refine the models; and the testing set is used to test the models' performance. 

\textbf{Re-sampling}.
Cross-validation is a popular re-sampling procedure employed to evaluate machine learning models on a limited number of data samples. The procedure has a parameter $k$, indicating the number of partitions that the given data is to be split into. Thus, this procedure is often called k-fold cross-validation. The samples of each partition are given the opportunity to be used as the validation set once and used to train the model k-1 times. The general procedure consists of the following steps:
\begin{enumerate}
    \item Shuffle the data randomly
    \item Split the data into k partitions
    \item For each unique partition: 
    \begin{enumerate}
        \item Take the partition as the validation set, and take the remaining $k-1$ partitions as a training set
        \item Fit a model on the training set and evaluate it on the validation set
        \item Retain the evaluation score and discard the model
    \end{enumerate}
    \item Summarize the performance of the model using the sample of model evaluation scores
\end{enumerate}

As a result, the model's validation score is the average of the K validation scores obtained. 



\chapter{Objectives}
\label{c:objective}

This section summarizes the research questions (RQs) and objectives of the thesis. Each formulated research question is accompanied with the findings from some related papers on specific aspects. We now detail each of them and elaborate on the objectives in response to these RQs.

\section{Creating a dataset}
\subsection{Research Question 1}
Fact-checking is the process of verifying information in text in order to determine its factuality and correctness. Presently, there is a lack of large real-life occurring fact-checking datasets for training and evaluating models. The first research question, therefore, explores the creation of a dataset for fact-checking.

\begin{description}
\item \textit{(RQ1) How can we create a large
number of real-life occurring claims and rich additional meta-information for training and evaluating models?}
\end{description}

This question is motivated and discussed in Paper 1, chapter \ref{chap:createdataset}. Researchers have questioned the credibility of information on the Web for more than a decade. There have been several datasets that focus on fact-checking or rumor detection. These datasets are often created by collecting the claims and labels from fact-checking websites, e.g., politifact.com or snopes.com. However, existing efforts either use small datasets consisting of real-life occurring claims or more massive datasets but consisting of artificially constructed claims such as FEVER. 

Therefore, we address this by creating the largest fact-checking dataset of real-life occurring claims collected from 26 fact-checking websites in English for claim fact-checking. The dataset, called MultiFC, consists of 34,918 claims, along with evidence pages, the context in which they occurred, and rich metadata. 

We also perform a thorough analysis to identify the dataset characteristics, such as entities mentioned in claims. We show the utility of the dataset by training state of the art fact-checking inference models and find that evidence pages and metadata significantly contribute to model performance. 



\section{Improving semantic representation} 
\label{s:semrepre}
\subsection{Research Question 2}
The second research question investigates semantic representation for NLP. We particularly focus on the compositionality detection of phrases as semantic indicators, which is meaningful for any semantic processing application, such as search engines.

\begin{description}
\item \textit{(RQ2) How can we detect the compositionality of a phrase contextually?}
\end{description}

Prior to our research, existing research treats phrases as either compositional or non-compositional in a deterministic manner. Motivated by this, we operationalize the compositionality of a phrase contextually rather than deterministic. For instance, \textit{heavy metal} could refer to a dense metal that is toxic, which is compositional, but it could also be non-compositional when it refers to a genre of music. Previous work acknowledges this property of compositionality theoretically (\cite{Lioma2017}), but no operational models implementing this have been presented to this day.

Given a multi-word phrase as input, we reason that the phrase is used in some narrative, e.g., a query, sentence, snippet, document, etc. We refer to this narrative as a \textit{usage scenario} of the phrase. We combine evidence extracted from this usage scenario of the phrase with the global context (frequently co-occurring terms) of the phrase and use this to enrich the word embedding representation of the phrase. 
We linearly combine the weights of the tokens that are obtained from the usage scenario and the global context. We further extend this representation with information extracted from external knowledge bases.



\section{Extending convolutional neural inference models} 
\label{s:extendcnn}
\subsection{Research Question 3}
This research question focuses on extending convolutional neural models for a particular NLP task. As a continuous study of using the fact-checking dataset (discussed in RQ1), we, therefore, investigate and extend the CNN model for fact-checking. 

\begin{description}
\item \textit{(RQ3) How can we infer the factuality of a claim as true, false, or half-true?}
\end{description}

In response to RQ3, we participated in the CLEF2018 fact-checking evaluation task with a proposed system. The task is described below: given a set of political debate claims that have been already identified as worth checking, the aim is for the system to determine whether the claim is likely to be true, half-true, false, or that it is unsure of its factuality.

As CLEF provides limited data (only 82 unique claims with labels), but the task of fact-checking relies on labeled data to train prediction models, finding suitable datasets for training is the first basic step. Furthermore, the task at
hand is more complex than traditional binary prediction (True/False) as graded truth values must be predicted, including the difficult “Half-True”. There are primarily three objectives that we take into consideration:

\begin{itemize}
    \item Select external claims with labels and a suitable proportion of samples. The suitable proportion here indicates we maintain a balance of class distribution with respect to the test dataset. 
    \item Retrieving the most relevant but suitable amount of external evidence (documents) for claims. The suitable amount here indicates we assume too few resources can lead to insufficiency, while too many external resources can lead to heavy noise.
    \item Find the best models and parameters and tune them to their best performance.
\end{itemize}

The three objectives are met by proceeding in a step-wise manner. Selecting external claims of high quality is the basis of the following steps. The multiple labels and their proportional samples have to be taken into account when selecting datasets with different labeling. Subsequently, retrieving the most relevant but adequate documents for these claims is significant to support the building of training models. Finally, selected features of documents should be fitted well into different models, of which the parameters should be tuned to improve the final results.

We leverage the challenge by characterizing our solutions as followings:
\begin{enumerate}
    \item We use step-wise modeling, instead of using a mixed model in the final step, i.e., we use traditional Bayes models for data prepossessing, including data selection (label mapping) and external source analysis (sufficiency analysis), and then build a CNN model based on the previous conclusion. The CNN model is composed of multi-scale kernels with different window sizes that learn from external sources to infer fact-checking labels
    \item We employ step-wise searching in retrieving supporting documents with as much of the original claim as possible while strategically retrieving enough documents, instead of just using keywords
\end{enumerate}

\subsection{Research Question 4}
This research question focus on extending the convolutional neural inference model for the relation extraction task. Relation extraction is an essential task in NLP to extract the relation between two entities in a given context, which is important for popularizing knowledge bases (KBs), e.g., DBpedia (\cite{auer2007dbpedia}),YAGO (\cite{kasneci2006yago}), and  Freebase (\cite{bollacker2008freebase}). 

\begin{description}
\item \textit{(RQ4) How can we enhance CNN represention for relation inference task?}
\end{description}

Prior to our work, most of the work in relation inference has been dominated by two lines of approaches, differentiated by the essence of the relation description: the kernel-based approaches (\cite{qian2008exploiting,mooney2006subsequence,bunescu2005shortest,zhang2006composite,nguyen2014employing,liu2013convolution}), and feature-based approaches (\cite{jiang2007systematic,kambhatla2004combining,nguyen2014employing,chan2010exploiting}). Feature-based methods rely on classification models that classify a collection of features, e.g., words, phrases, etc., to a relation directly. The feature-based approaches are computationally infeasible to generate features involving long-range dependencies. On the contrary, Kernel-based methods retain the original representation of features and use the features for kernel function to extract extra features between pairs of features. A kernel function is a similarity function satisfying certain properties. Specifically, a kernel function $K$ over the feature space $X$ is binary function $K$ satisfying: $X \times X \xrightarrow{} [0,\infty]$ mapping a pair of feature $x_1,x_2 \in X$ to their similarity score $K(x_1,x_2)$. 

There is a shared process in these approaches, i.e., to leverage linguistic analysis to map relation mentions into rich representations, which can be utilized by some statistical classifiers, including Maximum Entropy (\cite{nigam1999using,osborne2002using}) or Support Vector Machines (\cite{suykens1999least,scholkopf2001learning}). 

The linguistic analysis pipeline consists of several manually designed steps such as tokenization, chunking, part of speech tagging, parsing, and name tagging. However, heavy NLP feature engineering can lead to error propagation in relation extraction (\cite{mcclosky2010automatic,jurafsky2006proceedings,sun2016return}).

We address the problem by proposing a novel structural block-driven convolutional neural representation for relation extraction. The main advantage of the method is that we employ manually-determined features as opposed to manually selected features. To be specific, we detect the essential spans associated with entities through dependency relation analysis. We obtain those selective sequential tokens as blocks by gathering the parent, siblings, and children nodes of entities from the dependency tree. We then enhance the word encoding by enriching the word embedding with syntactic tag encoding (including the syntactic role and part-of-speech tags), and encoding the blocks with multi-scale CNNs. In addition, we add two extra inter-block representation with 1) a subtraction representation, i.e., the block vector subtracted by the two entity vectors, and 2) a multiplication representation, i.e., the product of the two entity vectors. Then, we fully concatenate the block-wise and inter-block wise representation to infer the resulting relation. 

This design has the advantages of being able to (1) eliminate the noise from irrelevant parts of a sentence; meanwhile, (2) enhance the relevant block representation with both block-wise and inter-block-wise semantically enriched representation. As a result, the encoding of a selective part of a sentence and the block's enhanced encoding lead to an improved effectiveness and efficiency. 


\section{Extending Self-Attention inference models} 
\label{s:extendsa}

In this section, we investigate extending self-attention representation from Transformer to improve its effectiveness. The extension is studied from different perspectives. We detail them from RQ5 to RQ7.


\subsection{Research Question 5}
This research question explores the extension of the self-attention model from the perspective of head roles to improve its effectiveness. 

\begin{description}
\item \textit{(RQ5) What is the impact of explicitly guiding attention heads on their resulting effectiveness?}
\end{description}

This research question is motivated by recent studies that have focused on developing approaches to understand how attention heads digest input texts, aiming to increase the interpretability of the model (\cite{clark2019does,voita2019analyzing,michel2019sixteen}). The findings of those analyses are aligned: while some attention heads of the Transformer often play linguistically-interpretable roles (\cite{clark2019does,voita2019analyzing}), others are found to be less critical and can be pruned without significantly impacting (indicating redundancy) or even improving (indicating potential errors contained in pruned heads) effectiveness (\cite{voita2019analyzing,michel2019sixteen}). Though the above studies demonstrate that the effectiveness of the attention heads is derived from different head roles, little prior work analyzes the impact of \textit{explicitly} adopting roles for the multiple heads.

To answer the research question, we define role-specific masks to guide the attention heads to attend to different parts of the input, such that different heads have their own attention computation ranges and are expected to play different roles. To achieve this, we first select important roles based on findings from recent studies on interpretable Transformers roles; then, we create masks with respect to those roles; and finally, the masks are incorporated into self-attention heads to guide the attention computation. 

For an input of length $n$, we define a role-specific mask as a n-by-n matrix where each element is either $-\infty$ (ignore) or 0 (include). Therefore, the mask can be incorporated such that a masked attention head (mh) can be expressed as:
\begin{equation}
\small
\label{eq:masked_att}
    \mathrm{mh(Q,K,V,M_r)=\softmaxname \left(\frac{QK^T+M_r}{\sqrt{d_k}}\right)V}
\end{equation}
where $M_r$ is a role-specific mask used to constrain the attention head; and the remaining is the same as discussed in section 2.2.3.

\subsection{Research Question 6}
This research question explores the extension of the self-attention model from the perspective of information aggregating to improve its effectiveness.


\begin{description}
\item \textit{(RQ6) What is the impact of aggregating attention heads with soft gates?}
\end{description}

A soft-gate indicates a unit that controls the contribution of information of a head via a product value between 0 to 1 rather than 0 or 1. This research question is motivated by the findings that the learned attention heads of Transformer have different importance, which is discussed in RQ 5. To address this, we propose to adopt a soft gating mechanism for the multi-head attention, such that the information of different heads is selectively ensembled in each layer of the multi-head attention. We achieve this by adapting the concatenating of heads into a weighted summation, enabling the gates to learn different attention heads' contributions.

\subsection{Research Question 7}
This research question investigates a novel inference model utilizing attention heads.

\begin{description}
\item \textit{(RQ7) How to encode a sentence in a hierarchical way rather than sequential?}
\end{description}

Prior to our work, most existing sentence representation approaches encode sentences as a whole, using the sequence of tokens or characters. These approaches fail to capture more complex or nuanced semantics of words in context because they use the linear order of words when capturing their semantics and do not take into account the hierarchical information of long-term dependency. For instance, "man" and "went" are long-term dependency in the sentence of "the man who wore a black jacket went into the room."  

To address this, we propose to use the hierarchical order of syntactic dependencies of words when capturing their semantics. We posit that this is better for modeling linguistic discontinuities, for example, "To his wife, Jim gave a fantastic present" where "To his wife" is fronted, incurring crossing lines in the tree structure of dependency. We thus present a novel sentence encoding approach, named MS Encoder (Major and Surrounding Encoder), which first uses syntactic dependency roles to divide the sentence into major and non-major segments, then encodes each segment separately, and finally aggregates the two-segment encodings through a multiplication alignment and weighted gates to create a joint representation of the sentence.

\section{Publications}

The thesis summarizes the work done during my Ph.D., starting November 1st 2017, and ending October 31st 2020. Therefore, it consists of a collection of papers published or submitted during my Ph.D., which fall into the three categories, as continuous studies. 

\textbf{Papers included in this thesis:}

[1] Isabelle Augenstein, Christina Lioma, Dongsheng Wang, Lucas Chaves Lima, Casper Hansen, Christian Hansen, and Jakob Grue Simonsen. MultiFC: A Real-World Multi-Domain Dataset for Evidence-Based fact-checking of Claims.  2019, Proceedings of the 2019 Conference on Empirical Methods in Natural Language Processing and the 9th International Joint Conference on Natural Language Processing (EMNLP-IJCNLP). Association for Computational Linguistics, p. 4684-4696

[2] Dongsheng Wang, Qiuch Li, Lucas Chaves Lima, Jakob Grue Simonsen and Christina Lioma. Contextual compositionality detection with external knowledge bases and word embeddings. The Web Conference 2019 - Companion of the World Wide Web Conference, WWW 2019. Association for Computing Machinery, p. 317-323 7 p, 2019.

[3] Dongsheng Wang, Jakob Grue Simonsen, Birger Larsen and  Christina Lioma. The Copenhagen team participation in the factuality task of the competition of automatic identification and verification of claims in political debates of the CLEF-2018 fact-checking Lab. CLEF 2018 Working Notes. Cappellato, L., Ferro, N., Nie, J-Y. \& Soulier, L. (eds.). 10 ed. CEUR-WS.org, 10 p. 98, 2018.

[4] Dongsheng Wang, Prayag Tiwari, Sahil Garg, Hongyin Zhu and Peter Bruza. Structural block driven enhanced convolutional neural representation for relation extraction, In : Applied Soft Computing Journal. 86, 9 p., 105913, 2020.

[5] Dongsheng Wang, Casper Hansen, Lucas Chaves Lima, Christian Hansen, Maria Maistro, Jakob Grue Simonsen and Christina Lioma. Multi-Head Self-Attention with Role-Guided Mask. ECIR2021.

[6] Dongsheng Wang, Lucas Chaves Lima, Casper Hansen, Maria Maistro, Jakob Grue Simonsen and Christina Lioma. Multi-Head Self-Attention with Weighted Gates. [in submission].

[7] Dongsheng Wang, Casper Hansen, Maria Maistro, Jakob Grue Simonsen and Christina Lioma. Encoding Major and Surrounding Sentence Segmentation Attentively and Jointly. [in submission].

\textbf{Papers published during the PhD but not included in the thesis:}

[1] Dongsheng Wang, Yongjuan Zhang, Zhengjun Wang, Tao Chen. Publishing E-RDF linked data for many agents by single third-party server. Semantic Technology: 7th Joint International Conference, JIST 2017, Gold Coast, QLD, Australia, 2017. 

[2] Hongyin Zhu, Yi zeng, Dongsheng Wang, Cunqing Huangfu. Species Classification for Neuroscience Literature Based on Span of Interest Using Sequence-to-Sequence Learning Model. Frontiers in Human Neuroscience, vol. 14, 128. 2020.

\chapter{Contributions}
\label{c:contribution}

The main contributions of the thesis are summarized as follows:

\textbf{Creating dataset:}
\begin{enumerate}
    \item We provided the largest fact-checking dataset of real-life occurring claims for the purpose of automatic verification, which we make publicly available to the community. (chapter \ref{chap:createdataset})
    \item We provided a benchmarking dataset of contextualized compositionality detection that we make publicly available to the community. (chapter \ref{chap:comp_detection})
\end{enumerate}

\textbf{Semantic Representation:}
\begin{enumerate}
    \item We proposed a novel model that detects phrase compositionality under different contexts and that outperforms the state of the art performance in the area. (chapter \ref{chap:comp_detection})

\end{enumerate}

\textbf{Convolutional Neural based Inference Models:}
\begin{enumerate}
    \item We introduced a novel inference model based on multi-scale CNNs, with which we achieved the best performance in the CLEF2018 fact-checking competition. (chapter \ref{chap:fact_checking})
    \item We proposed a novel relation extraction approach based on a convolutional neural inference model with few features called the structural block-driven convolutional neural model. The model achieved new state-of-the-art performance in the KBP37 dataset. (chapter \ref{chap:structblock})
\end{enumerate}

\textbf{Self-Attention based Inference Models:}
\begin{enumerate}
    \item We proposed Transformer-Guided-Attn, an extended method to explicitly guide the attention heads of the Transformer using role-specific masks. We demonstrate that incorporating multiple role masks into multi-head attention can consistently improve performance on both classification and machine translation tasks. (chapter \ref{chap:guidedheads})
    \item We proposed Transformer-Soft-Gated, a method that selectively aggregates the multi-head attention at each layer. We demonstrate that incorporating a weighted gating layer into multi-head attention can improve performance on both linguistic probing tasks and machine translation tasks. (chapter \ref{chap:gatedheads})
    \item We proposed the MS Encoder, a method that first uses syntactic dependency roles to divide the sentence into major and non-major segments, then encodes each segment separately, and finally aggregates the two segment encodings with a multiplication alignment and weighted gates to create a joint representation of the sentence. We demonstrate that MS encoder improves performance on both classification, paired-sentence inference, and machine translation tasks. (chapter \ref{chap:msencoder})
    
    
\end{enumerate}

\chapter{Future Work}
\label{c:futurewrk}
There remain some interesting open works that we can explore. 





\textbf{Future of inference learning for NLP}. Here are personal thoughts. Broadly speaking, a deep learning model is a geometric transformation between one vector space to another. As a result, most of what deep learning is skilled at is to output either a predefined category or a sequence of items (from a seq2seq model). Even if the model can produce an abstraction for a document; the model still does not "understand" the abstract's content. Therefore, the deep learning models can hardly support such tasks that require the outputs to contain some reasoning information, e.g., programming data structure algorithms. To address this, a slightly more structured and sophisticated output can be designed for researchers to enhance the inference learning model from a different viewpoint. It is worth to mention the researchers of symbolisms (e.g., ontology, description logics, expert systems, knowledge base, etc.) have worked in this field for decades; however, they could not have genuinely thrived till today. As a high-level future work, new forms of learning should be designed for NLP. One potential solution can be designing new tasks that require output with reasoning information, somehow challenging but potentially achievable forms, followed by developing new models for that.

\chapter{Conclusions}
\label{c:conclusion}

This thesis focuses on semantic representation and inference learning for NLP. We have presented the contributions in the three areas: creating datasets, improving semantic representation, and extending inference learning models. 

In terms of creating dataset, we have contributed two datasets in fact-checking and compositionality detection, respectively. Precisely, we have provided the largest fact-checking dataset of real-life occurring claims for automatic factuality verification, which is useful for training and evaluating machine learning and NLP models. Also, we offered a benchmarking dataset of contextual compositionality detection to the community. This dataset enables researchers to evaluate their compositionality detection model contextually based on usage scenarios rather than in a deterministic manner. 

For semantic representation, we have presented a novel compositionality detection model for phrase semantic indicator. Specifically, the model can contextually identify the phrase compositionality in specific usage scenarios. We posit this is essential for capturing semantic indicators such as in queries, sentences, etc. 

For inference learning, we have extended convolutional neural models and transformer self-attention models, respectively. 

For convolutional neural models, we have presented two approaches of extending convolutional neural models in fact-checking and relation extraction tasks, respectively. Precisely, we brought multi-scale CNNs that extend the CNN model in fact-checking, which strongly outperforms the other CLEF2018 fact-checking competition models. We also introduced a structural block-driven convolutional neural inference model that employs CNN to encode the blocks with enhanced word encoding (word embedding and syntactic tag embedding). We demonstrated that the enhanced encoding of blocks (selective parts from texts) could improve relation inference efficiency and effectiveness.

For the self-attention model, we demonstrated that the transformer's self-attention could be extended in various ways to improve its effectiveness. First, we presented that self-attention with guided roles holds the promise to provide an interpretable and improved inference model. Also, when we aggregate the multi-head attention with different mechanisms, such as a weighted gating layer, we can obtain improvements from the same attention heads trained on NLP tasks. Besides, we presented the MS Encoder, a method that uses syntactic dependency to divide the sentence into major and non-major segments and hierarchically encode them. We posit that this is better for modeling linguistic discontinuities.

\part{Creating Training Dataset}\label{part:II}
\chapter{MultiFC: A Real-World Multi-Domain Dataset for Evidence-Based Fact Checking of Claims
} \label{chap:createdataset}
The work introduced in this chapter is based on a paper that have been published as:

[1] Isabelle Augenstein, Christina Lioma, Dongsheng Wang, Lucas Chaves Lima, Casper Hansen, Christian Hansen, \& Jakob Grue Simonsen. MultiFC: A Real-World Multi-Domain Dataset for Evidence-Based Fact Checking of Claims.  2019, Proceedings of the 2019 Conference on Empirical Methods in Natural Language Processing and the 9th International Joint Conference on Natural Language Processing (EMNLP-IJCNLP). Association for Computational Linguistics, p. 4684-4696

\includepdf[pages={2,3,4,5,6,7,8,9,10,11,12}]{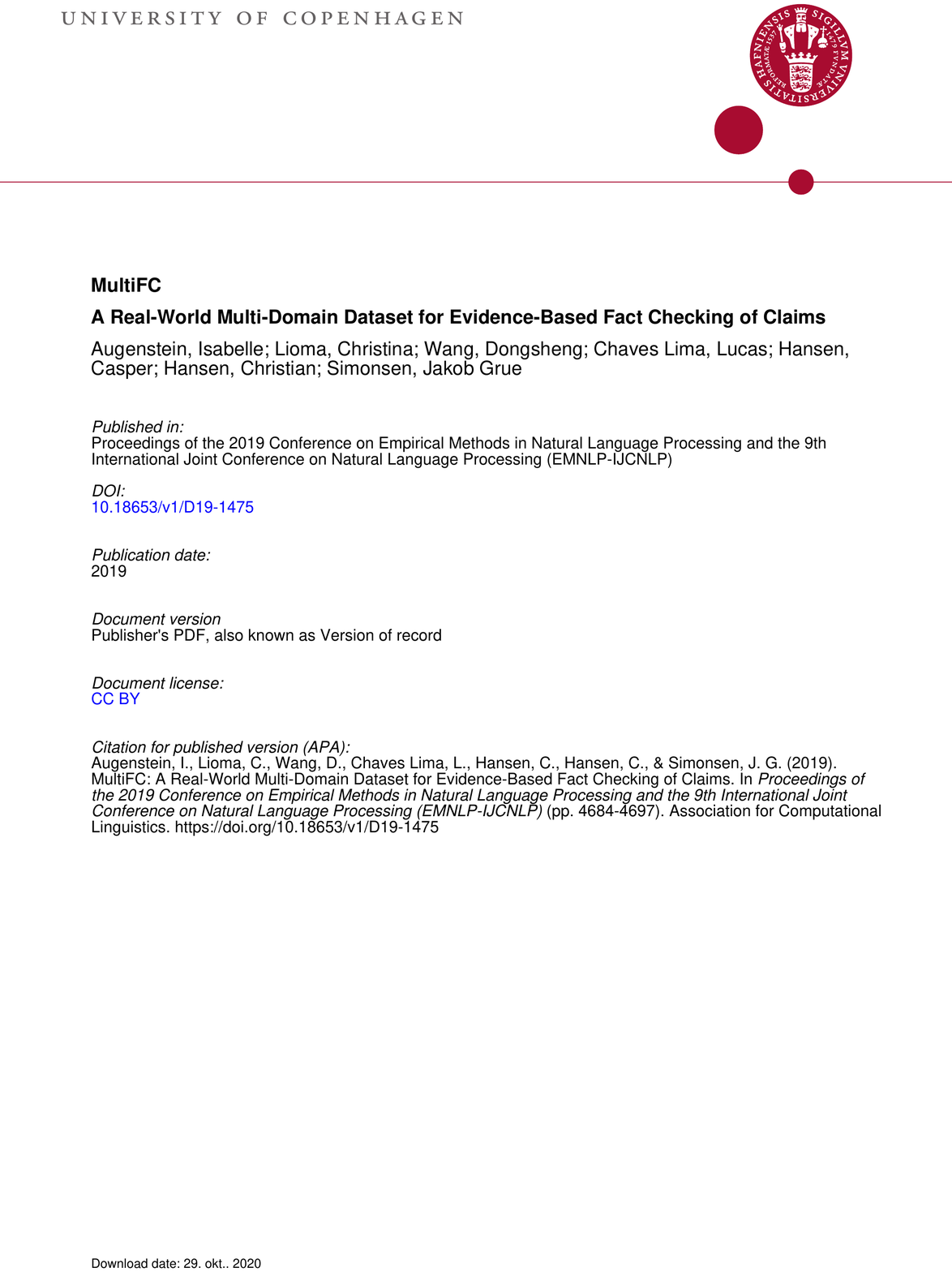}

\part{Semantic Representation}\label{part:III}
\chapter{Contextual compositionality detection with external knowledge bases and word embeddings}
\label{chap:comp_detection}

This work presented in this chapter is based on a published paper:

[2] Dongsheng Wang, Qiuch Li, Lucas Chaves Lima, Jakob Grue Simonsen, Christina Lioma. Contextual compositionality detection with external knowledge bases and word embeddings. The Web Conference 2019 - Companion of the World Wide Web Conference, WWW 2019. Association for Computing Machinery, p. 317-323 7 p, 2019.

\includepdf[pages={2,3,4,5,6,7,8}]{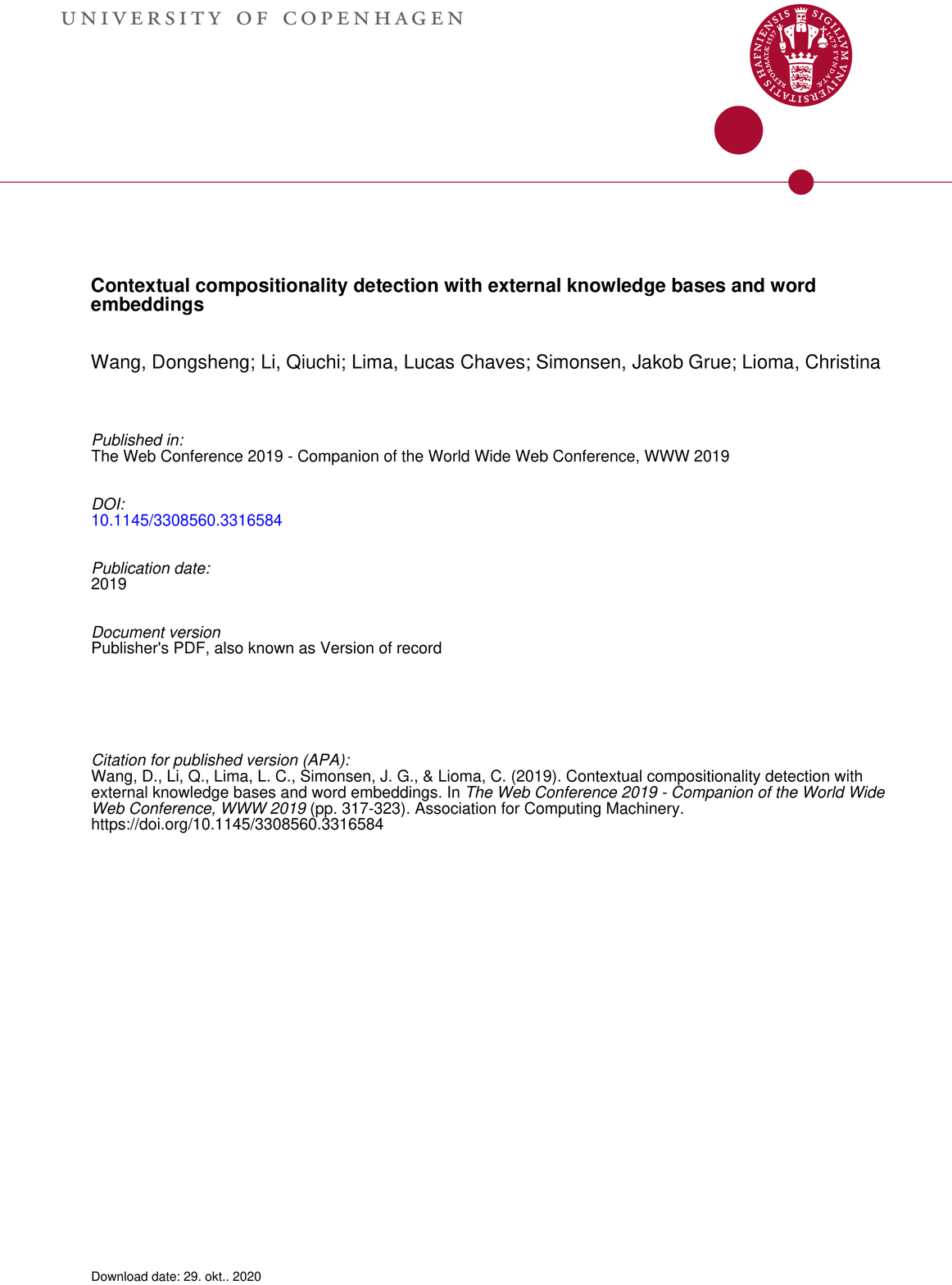}

\part{Convolutional Neural based Inference Learning}\label{part:IV}
\chapter{The Copenhagen team participation in the factuality task of the competition of automatic identification and verification of claims in political debates of the CLEF-2018 Fact Checking Lab}\label{chap:fact_checking}

The work presented in this chapter is based on a published work, which is the a competition of fact checking CLEF2018.

[3] Dongsheng Wang, Jakob Grue Simonsen, B. Larsen \& Christina Lioma. The Copenhagen team participation in the factuality task of the competition of automatic identification and verification of claims in political debates of the CLEF-2018 Fact Checking Lab. CLEF 2018 Working Notes. Cappellato, L., Ferro, N., Nie, J-Y. \& Soulier, L. (eds.). 10 ed. CEUR-WS.org, 10 p. 98, 2018.

\includepdf[pages={1,2,3,4,5,6,7,8,9,10,11}]{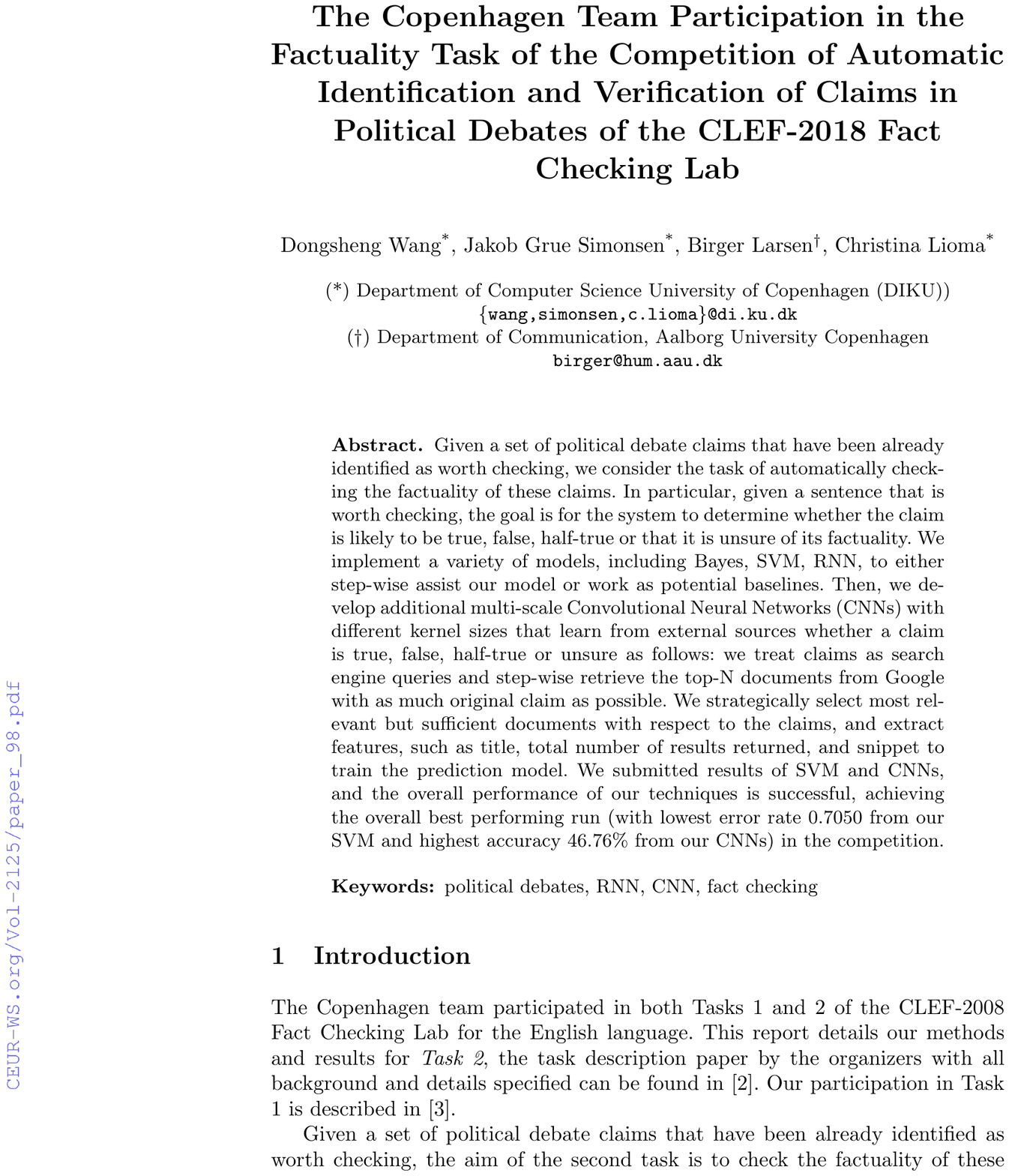} 
\chapter{Structural block driven enhanced convolutional neural representation for relation extraction}
\label{chap:structblock}

This work presented in this chapter is baesd on a published journal:

[4] Structural block driven enhanced convolutional neural representation for relation extraction. Wang, Dongsheng, Tiwari, P., Garg, S., Zhu, H. \& Bruza, P., 2020, In : Applied Soft Computing Journal. 86, 9 p., 105913.

\includepdf[pages={1,2,3,4,5,6,7,8,9}]{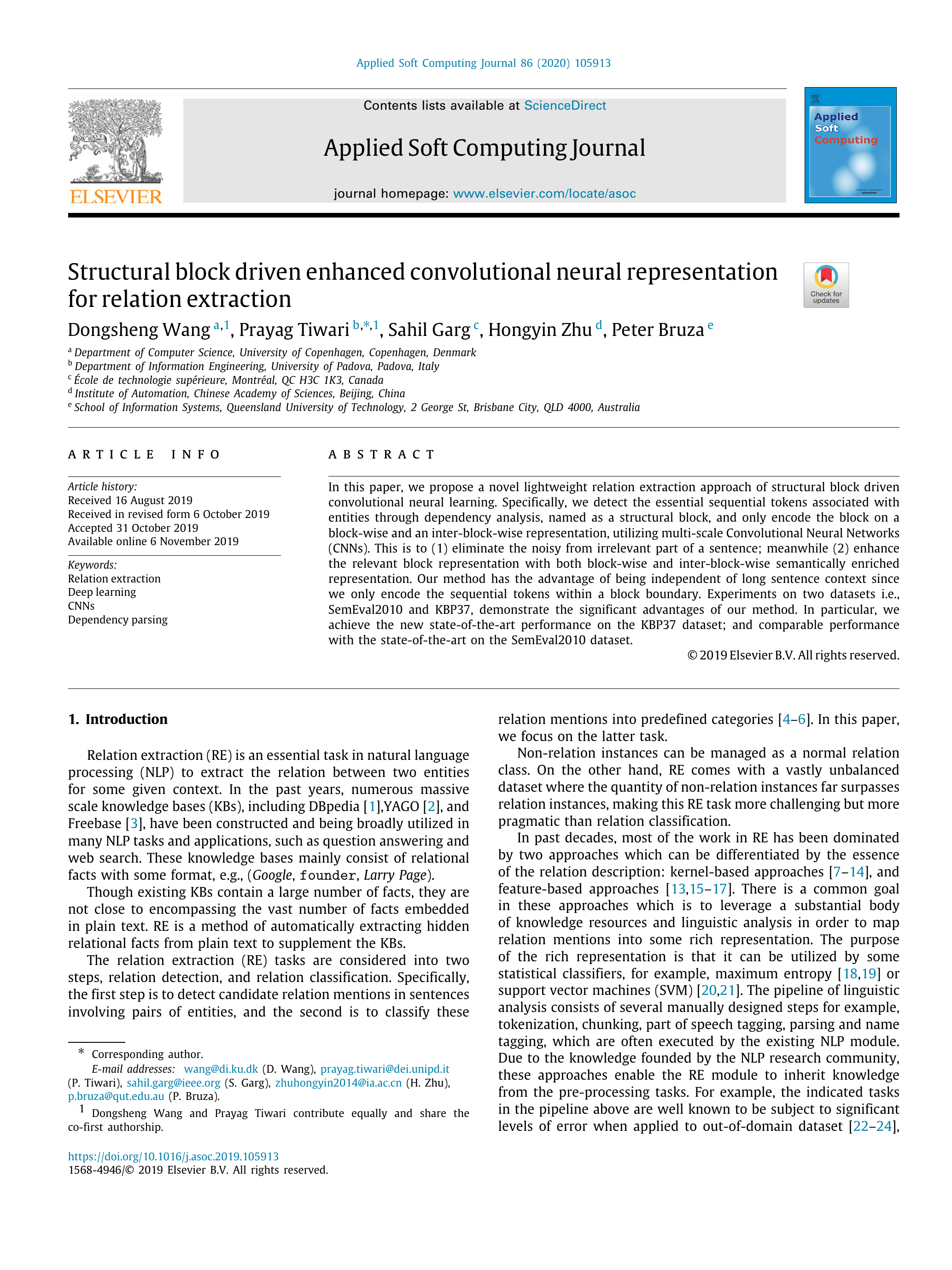} 

\part{Self-Attention based Inference Learning}\label{part:V}
\chapter{Multi-Head Self-Attention with Role-Guided Mask}
\label{chap:guidedheads}

This work presented in this chapter is based on a published work:

[5] Dongsheng Wang, Casper Hansen, Lucas Chaves Lima, Christian Hansen, Maria Maistro, Jakob Grue Simonsen, Christina Lioma. Multi-Head Self-Attention with Role-Guided Mask. ECIR2021. Lecture Notes in Computer Science, vol 12657. Springer, Cham. https://doi.org/10.1007/978-3-030-72240-1\_45

\includepdf[pages={1,2,3,4,5,6,7,8}]{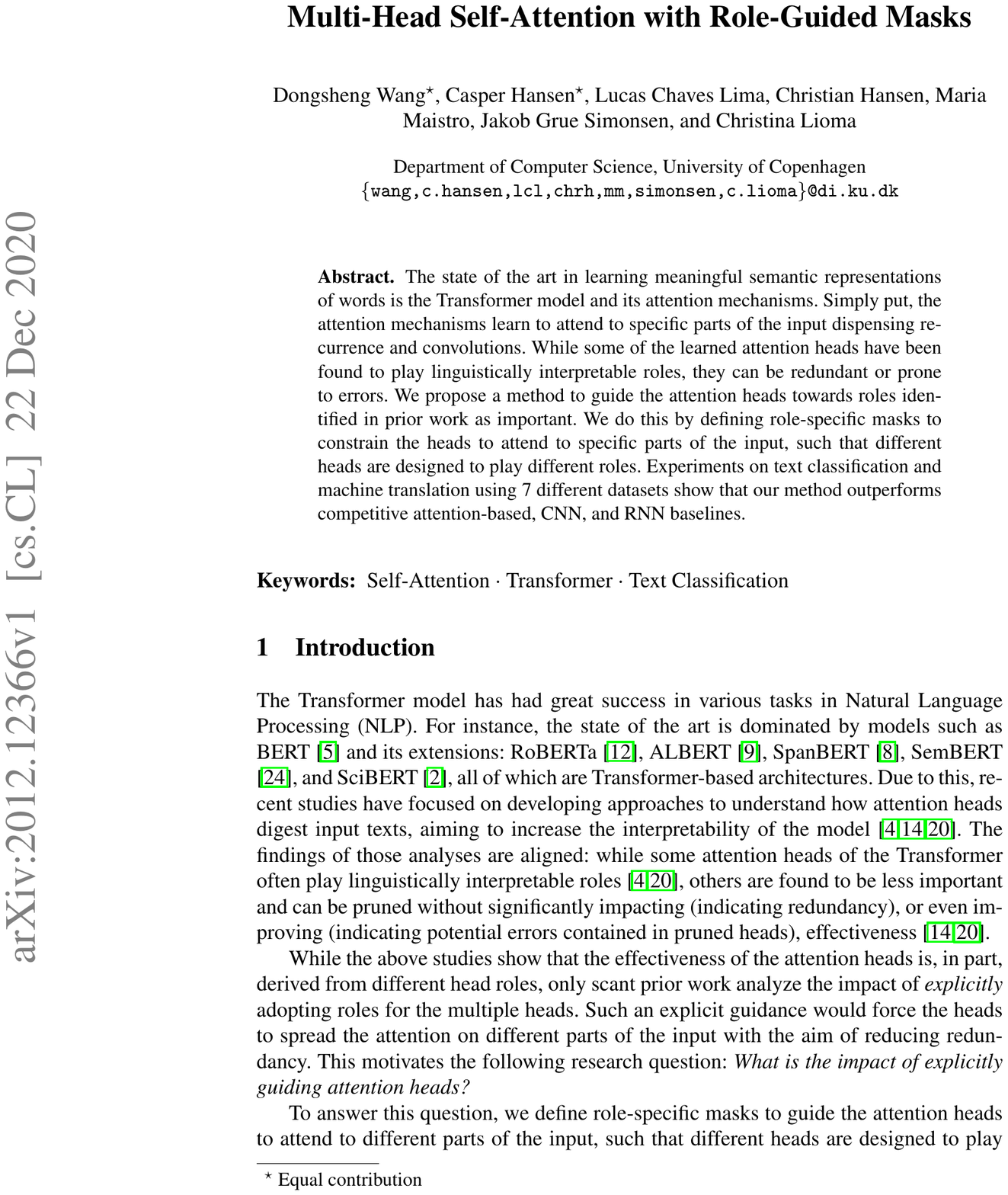}
\chapter{ Multi-Head Self-Attention with Weighted Gates}
\label{chap:gatedheads}

The work presented in this chapter is based on a paper that has been submitted:

[6] Dongsheng Wang, Lucas Chaves Lima, Casper Hansen, Maria Maistro, Jakob Grue Simonsen, Christina Lioma. Multi-Head Self-Attention with Weighted Gates. ECIR2021 [submitted].

\includepdf[pages={1,2,3,4,5,6,7,8}]{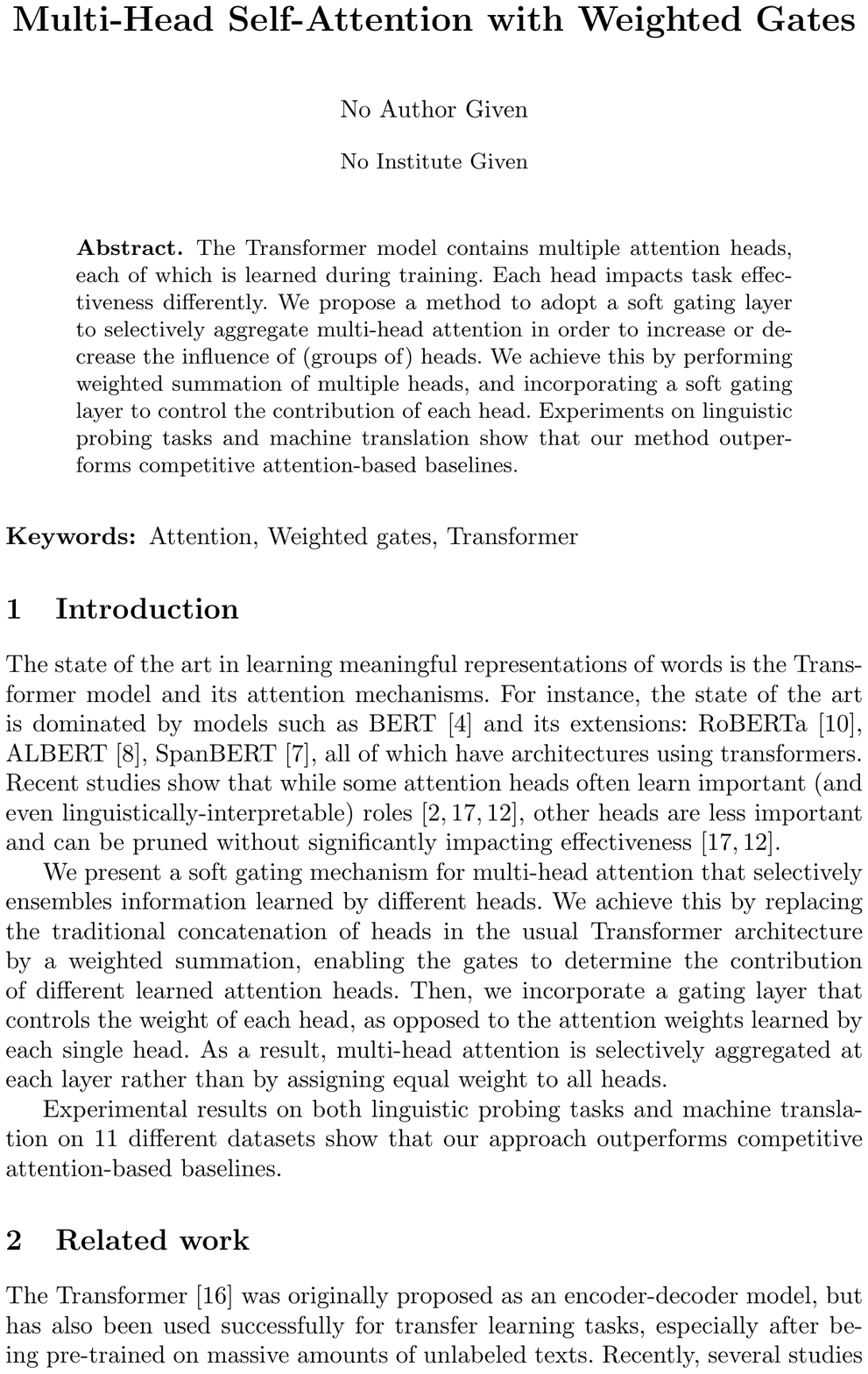} 
\chapter{Encoding Major and Surrounding Sentence Segmentation Attentively and Jointly}
\label{chap:msencoder}

The work presented in this chapter is based on a paper that has been submitted:

[7] Dongsheng Wang, Casper Hansen, Maria Maistro, Jakob Grue Simonsen, Christina Lioma. Encoding Major and Surrounding Sentence Segmentation Attentively and Jointly. AAAI2021 [submitted].

\includepdf[pages={1,2,3,4,5,6,7,8,9}]{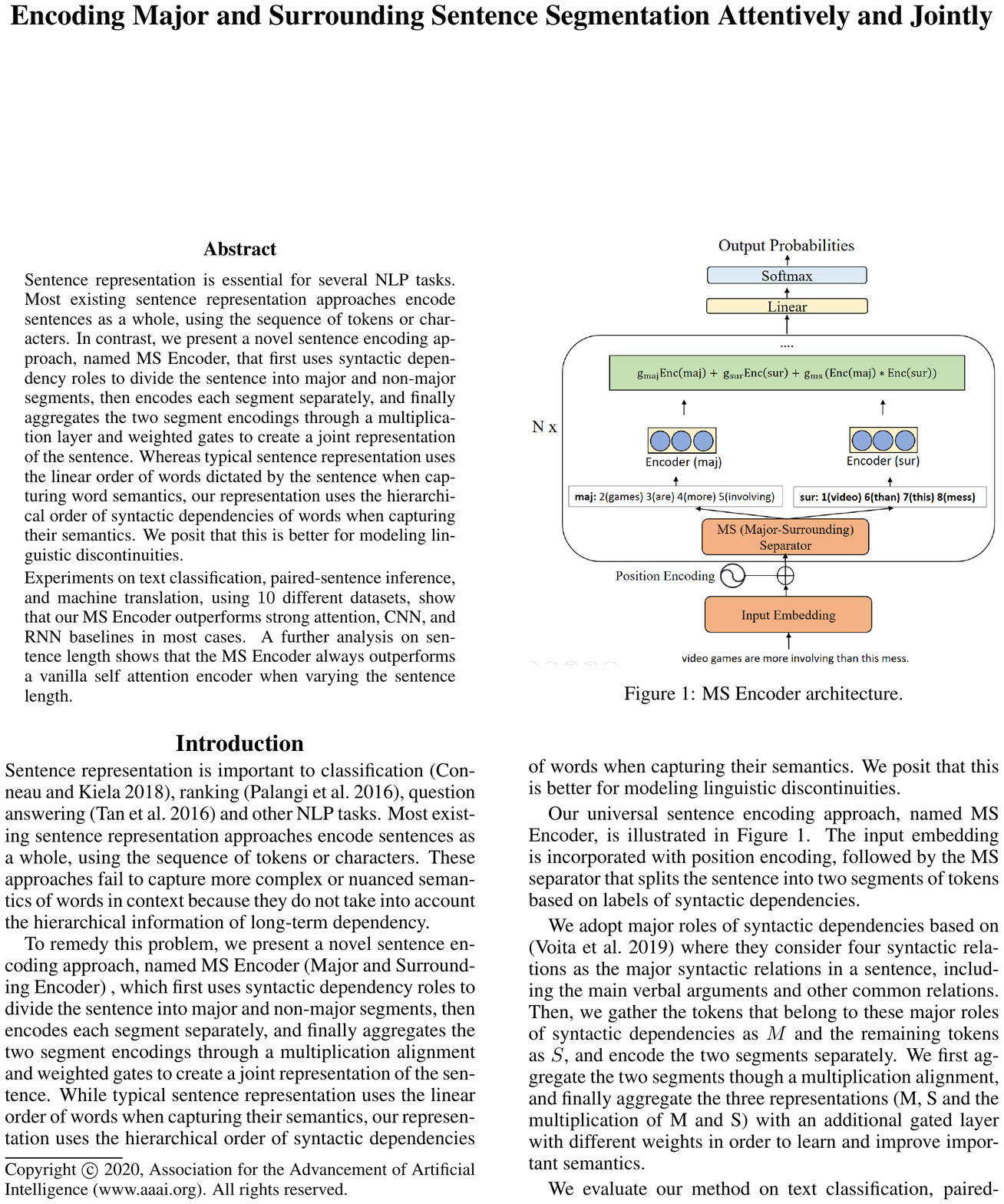}



\medskip

\printbibliography[
heading=bibintoc,
title={Bibliography}]

\clearpage

\end{document}